\documentclass[10pt, a4paper]{article}
\pdfoutput=1
\usepackage{lrec}
\usepackage{multibib}
\newcites{languageresource}{Language Resources}
\usepackage{graphicx}
\usepackage{tabularx}
\usepackage{caption}
\usepackage{array}
\usepackage[normalem]{ulem}
\usepackage{paralist}
\usepackage{booktabs}

\usepackage{hyperref}
\usepackage{xstring}
\usepackage{bold-extra}

\usepackage{color}

\usepackage{mathptmx}
\usepackage[T2A, T1]{fontenc}
\usepackage[russian, english]{babel}
\usepackage[utf8]{inputenc}

\newcommand{\secref}[1]{\StrSubstitute{\getrefnumber{#1}}{.}{}}

\title{
Prague Dependency Treebank - Consolidated 1.0 
}

\name{\parbox{\textwidth}{\centering Jan Hajič, Eduard Bejček, Jaroslava Hlaváčová, Marie Mikulová,  Milan Straka, \\ Jan Štěpánek, Barbora Štěpánková}}

\address{Charles University, Faculty of Mathematics and Physics, \\
		Institute of Formal and Applied Linguistics              \\
		Malostranské náměstí 25, 118 00 Prague 1, Czech Republic \\
		\texttt{\{hajic,hlavacova,mikulova,straka,stepankova\}@ufal.mff.cuni.cz}         \\
}

\abstract{
We present a richly annotated and genre-diversified language resource, the Prague Dependency Treebank-Consolidated 1.0 (PDT-C 1.0), the purpose of which is - as it always been the case for the family of the Prague Dependency Treebanks - to serve both as a training data for various types of NLP tasks as well as for linguistically-oriented research. PDT-C 1.0 contains four different datasets of Czech, uniformly annotated using the standard PDT scheme (albeit not everything is annotated manually, as we describe in detail here). The texts come from different sources: daily newspaper articles, Czech translation of the Wall Street Journal, transcribed dialogs and a small amount of user-generated, short, often non-standard language segments typed into a web translator. Altogether, the treebank contains around 180,000 sentences with their morphological, surface and deep syntactic annotation. The diversity of the texts and annotations should serve well the NLP applications as well as it is an invaluable resource for linguistic research, including comparative studies regarding texts of different genres. The corpus is publicly and freely available.
\\ \newline \Keywords{language resource, textual corpus, treebank, morphology, syntax, semantics, lexicon }
}

\begin{document}

\maketitleabstract

\section{Introduction}
In this paper, we present a richly annotated and genre-diversified language resource, the Prague Dependency Treebank-Consolidated version 1.0 (PDT-C in the sequel). PDT-C \citelanguageresource{lrPDT-C}\footnote{\url{https://ufal.mff.cuni.cz/pdt-c}} is a treebank from the family of PDT-style corpora developed in Prague (for more information, see \newcite{haj17}). The main features of this annotation style are:
\begin{itemize}
\item it is based on a well-developed dependency syntax theory known as the Functional Generative Description \cite{SgallHP:1986},
\item interlinked hierarchical layers of annotation,
\item deep syntactic layer with certain semantic features.
\end{itemize}

From 2001, when the first PDT-treebank was published, various branches of PDT-style corpora have been developed with different volumes of manual annotation on varied types of texts, differing in both the original language and genre specification. The treebanks were built with different intention. Manual annotation included in them covered a certain part of the PDT-annotation scheme (see Sect.~\secref{sec:from}). 

In the PDT-C project, we integrate four genre-diversified PDT-style corpora of Czech texts (see Sect.~\secref{sec:data}) into one consolidated edition uniformly and manually annotated at the morphological, surface and deep syntactic layers (see Sect.~\secref{sec:layers}). In the current PDT-C 1.0 release, manual annotation has been fully performed at the lowest morphological layer (lemmatization and tagging); also, basic phenomena of the annotation at the highest deep syntactic layer (structure, functions, valency) have been done manually in all four datasets.\footnote{Consolidation and fully manual annotation of the surface-syntactic layer is planned for the next, 2.0 version of PDT-C.} 

The difference from the separately published original treebanks can be briefly described as follows:

\begin{itemize}
\item it is published in one package, to allow easier data handling for all the datasets;
\item the data is enhanced with a manual linguistic annotation at the morphological layer and new version of morphological dictionary is enclosed (see Sect.~\secref{sec:morph});
\item a common valency lexicon for all four original parts is enclosed;
\item a number of errors found during the process of manual morphological annotation has been corrected (see also Sect.~\secref{sec:morph}).
\end{itemize}

PDT-C 1.0 is provided as a digital open resource accessible to all users via the LINDAT/CLARIN repository.\footnote{{\url{http://hdl.handle.net/11234/1-3185}}}

\begin{table*}[t]
\captionsetup{justification=centering}
\begin{center}
\begin{tabular}{lllll}
Dataset/Type of annotation & Written & Translated & Spoken & User-generated
\\\hline\hline
Audio  & non-applicable & non-applicable & provided &
non-applicable
\\
ASR transcript & non-applicable & non-applicable & provided &
non-applicable
\\
Transcript & non-applicable & non-applicable & manually &
non-applicable
\\
\multicolumn{5}{c}{Morphological layer\vbox{\vskip 1.5em}}
\\\hline
Speech reconstruction & non-applicable & non-applicable & manually &
non-applicable
\\
\textbf{Lemmatization} & manually & \textbf{manually} & \textbf{manually} & \textbf{manually}
\\
\textbf{Tagging} & manually & \textbf{manually} & \textbf{manually} & \textbf{manually}
\\
\multicolumn{5}{c}{Surface syntactic layer\vbox{\vskip 1.5em}}
\\\hline
Dependency structure & manually & automatically & automatically &
automatically
\\
Surface syntactic functions & manually & automatically & automatically &
automatically
\\
Clause segmentation & manually & not annotated & not annotated & not
annotated
\\
\multicolumn{5}{c}{Deep syntactic layer\vbox{\vskip 1.5em}}
\\\hline
Deep syntactic structure & manually & manually & manually & manually
\\
Deep syntactic functions & manually & manually & manually & manually
\\
Verbal valency & manually & manually & manually & manually
\\
Nominal valency & manually & not annotated & not annotated & not
annotated
\\
Grammatemes & manually & not annotated & not annotated & not annotated
\\
Coreference & manually & manually & manually & not annotated
\\
Topic-focus articulation & manually & not annotated & not annotated &
not annotated
\\
Bridging relations & manually & not annotated & not annotated & not
annotated
\\
Discourse & manually & not annotated & not annotated & not annotated
\\
Genre specification & manually & not annotated & not annotated & not
annotated
\\
Quotation & manually & not annotated & not annotated & not
annotated
\\
Multiword expressions & manually & not annotated & not annotated & not
annotated\\
\end{tabular}
\caption{Overview of various types of annotation and their realization
  in the datasets (new manual annotation made to PDT-C 1.0 is indicated in bold)}
\label{tab:annot}
\end{center}
\end{table*}

\section{Related Work}
\label{sec:related}
There is a wide range of corpora with rich linguistic annotation, e.g., Penn Treebank \cite{marcus1993building}, its successors PropBank \cite{kingsbury2002treebank} and NomBank \cite{meyers2004annotating} and OntoNotes \cite{Hovy:2006:O9S:1614049.1614064}; for German, there is Tiger \cite{brants2002tiger} and Salsa \cite{burchardt2006salsa}. 

The Prague Dependency Treebank is an effort inspired by the PennTreebank \cite{marcus1993building} (but annotated natively in dependency-style) and it is unique in its attempt to systematically include and link different layers of language including the deep syntactic layer. The PDT project has been successfully developed over the years and  PDT-annotation scheme has been used for other in-house development of related treebanks of Czech texts (see next Sect.~\secref{sec:from}) and also for treebanks originating elsewhere: HamleDT \cite{biblio:ZeMaHamleDTTo2012}, Slovene Dependency Treebank \cite{dvzeroski2006towards}, Greek Dependency Treebank \cite{prokopidis2005theoretical}, Croatian Dependency Treebank \cite{berovic2012croatian}, Latin Dependency Treebank \cite{bamman2006design}, and Slovak National Corpus \cite{vsimkova2006}. 

\section{From PDT to PDT-C}
\label{sec:from}
The first version of the Prague Dependency Treebank (PDT in the sequel) was published in 2001 \citelanguageresource{biblio:HaViPragueDependency2001}. It only contained annotation at the morphological and surface syntactic layers, and a very small “preview” of how the deep syntactic annotation might look like. The full manual annotation at all three annotation layers including the deep syntactic one is present in the second version, PDT 2.0, published in 2006 \citelanguageresource{biblio:HaPaPragueDependency2006}. The later versions of PDT did not bring more annotated data, but enriched and corrected the annotation of PDT 2.0 data. The latest edition of the core PDT corpus is version 3.5 \citelanguageresource{lrPDT35}\footnote{\url{http://ufal.mff.cuni.cz/pdt3.5}} encompassing all previous versions, corrections and additional annotation made under various projects between 2006 and 2018 on the original texts (clause segmentation at the surface syntactic layer, annotation of bridging relations, discourse, genre specifications, etc.; see Sect.~\secref{sec:layers}).

A slightly modified (simplified) scenario has been used for other PDT-corpora based on Czech texts: Prague Czech-English Dependency Treebank (the latest published versions is PCEDT 2.0 \citelanguageresource{PCEDT-LR}\footnote{\url{https://ufal.mff.cuni.cz/pcedt2.0}} and  PCEDT 2.0 Coref \citelanguageresource{PCEDTcoref}\footnote{\url{https://ufal.mff.cuni.cz/pcedt2.0-coref}}), Prague Dependency Treebank of Spoken Czech (the latest published version is 2.0 \citelanguageresource{pdtsc-LR}\footnote{\url{https://ufal.mff.cuni.cz/pdtsc2.0}}), and unpublished small treebank PDT-Faust.\footnote{\url{https://ufal.mff.cuni.cz/grants/faust}}  In contrast to the original project of PDT, in these treebanks, the morphological and surface syntactic annotations were done automatically, and the manual annotation at the deep syntactic layer is simplified: grammatemes, topic-focus articulation, nominal valency, and other special annotations are absent.

In the PDT-C project, we aim to provide all these included treebanks with full manual annotation at the lower layers and unify and correct annotation at all layers. Specifically, the data in PDT-C 1.0 is (mainly) enhanced with a manual annotation at the morphological layer, consistently across all the four original treebanks (see Sect.~\secref{sec:morph}).

\begin{table*}[t]
\centering
\begin{tabular}{l|rrrrr}
                     & Written   & Translated & Spoken   & Internet & Total
\\\hline
Morphological layer  & 1,957,247 &  \textbf{1,162,072} &  \textbf{742,257} & \textbf{33,772} & 3,895,348
\\
Surface syntactic layer & 1,503,739 &  1,162,072 &  742,257 & 33,772 & 3,441,840
\\
Deep syntactic layer    &   833,195 &  1,162,072 &  742,257 & 33,772 & 2,771,296
\\
\end{tabular}
\caption{Volume of the datasets (number of tokens, new manual annotation made to PDT-C 1.0 is indicated in bold)}
\label{tab:volume}
\end{table*}

\section{Data}
\label{sec:data}
As mentioned above, PDT-C 1.0 consists of four different datasets coming from PDT-corpora of Czech published earlier: dataset of written texts, dataset of translated texts, dataset of spoken texts, datasets of user-generated texts. These datasets are described in the following subsections.

The data volume is given in Table~\ref{tab:volume}. Altogether, the consolidated treebank contains 3,895,348 tokens with manual morphological annotation and 2,771,296 tokens with manual deep syntactic annotation (manual annotation of the surface syntactic layer is contained only in the dataset of written texts and it consists of 1,503,739 tokens). Table~\ref{tab:annot} presents an overview of various types of annotation at the three annotation layers (see Sect.~\secref{sec:layers}) in each dataset and the information of the manner in which the annotations was carried out. The newly provided manual morphological annotation made to PDT-C 1.0 is indicated in bold.

The markup used in PDT-C 1.0 is the language-independent Prague Markup Language (PML), which is an XML subset (using a specific scheme) customized for multi-layered linguistic annotation \cite{biblio:PaStRecentAdvances2008}.

\subsection{Written Data}
The dataset of written texts is based on the core \textbf{Prague Dependency Treebank} \citelanguageresource{lrPDT35}. The data consist of articles from Czech daily newspapers and magazines.

The annotation in the PDT dataset is the richest one and it has completely been perfomed manually (cf. Table~\ref{tab:annot}). In PDT-C 1.0, the annotation at the morphological layer has been checked and corrected to reflect the updated morphological annotation guidelines and also to be fully consistent with the morphological dictionary (see Sect.~\secref{sec:morph}).

\subsection{Translated Data}
The dataset of translated texts comes from the \textbf{Prague Czech-English Dependency Treebank} (PCEDT in the sequel \citelanguageresource{PCEDTcoref}). PCEDT is a (partially) manually annotated Czech-English parallel corpus. The English part consists of the Wall Street Journal sections of the Penn Treebank \citelanguageresource{penntb-LR}, with the original annotation preserved, but also converted to the PDT-style morphology and dependency syntax annotation. A manually annotated deep syntactic layer was added. The Czech part of PCEDT used in the PDT-C consolidated edition, has been manually (and professionally, with multiple quality control passes) translated from the English original, sentence to sentence \cite{biblio:HaHaAnnouncingPrague2012}. 

In the PDT-C 1.0 translation dataset coming from PCEDT, there is a simplified manual annotation at the deep syntactic layer. The annotation at surface syntactic layer is still done only by an automatic parser and there is the new manual annotation at the morphological layer (cf. Table~\ref{tab:annot}).

\subsection{Spoken Data}
The dataset of spoken texts is taken from the \textbf{Prague Dependency Treebank of Spoken Czech} (PDTSC in sequel \citelanguageresource{pdtsc-LR}). PDTSC contains slightly moderated testimonies of Holocaust survivors from the Shoa Foundation Visual History Archive\footnote{\url{https://ufal.mff.cuni.cz/cvhm/vha-info.html}} and dialogues (two participants chat over a collection of photographs) recorded for the EC-funded Companions project.\footnote{\url{http://cordis.europa.eu/project/rcn/96289\_en.html}}

The spoken data differs from the other included PDT-corpora mainly in the “spoken” part of the corpus \cite{biblio:MiMiPDTSC202017}. The process starts at the “audio” layer, which contains the audio signal. The next layer contains the transcript as produced by an automatic speech recognition engine (also coming from the Companions project). The word layer contains manual transcription of the recorded speech, and the morphological layer contains the reconstructed, i.e. grammatically corrected version of the sentence. From this point on, annotation on the upper layers is standard (see Sect.~\secref{sec:layers}). 

In the PDT-C 1.0 spoken dataset coming from PDTSC, there is also only a simplified manual annotation at the deep syntactic layer, the annotation at surface syntactic layer is still done only automatically and there is the new manual annotation at the morphological layer (cf. Table~\ref{tab:annot}).

\subsection{User-generated Data}
The dataset of user-generated texts comes from the \textbf{PDT-Faust} corpus. PDT-Faust is a small treebank containing short segments (very often with non-standard as well as expressive, obscene and/or vulgar content) typed in by various users on the \url{reverso.net} webpage for translation.

In the PDT-C 1.0 user-generated content dataset, there is also only a simplified manual annotation at the deep syntactic layer. Compared to the other datasets, there is no annotation of coreference. The annotation at surface syntactic layer is performed automatically and there is the added manual annotation at the morphological layer (cf. Table~\ref{tab:annot}).

\begin{figure}[ht!]
\captionsetup{justification=centering}
\begin{center}
\includegraphics[width=0.40\textwidth]{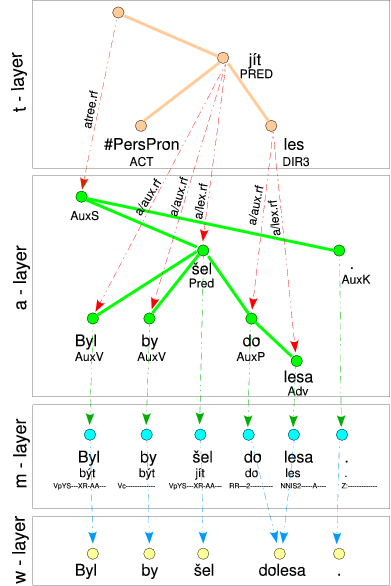}
\caption{ Linking the layers in PDT-C 1.0}
\label{fig-layers}
\end{center}
\end{figure}

\begin{figure}[ht!]
\captionsetup{justification=centering}
\begin{center}
\hspace*{-0.07in}\includegraphics[width=0.40\textwidth]{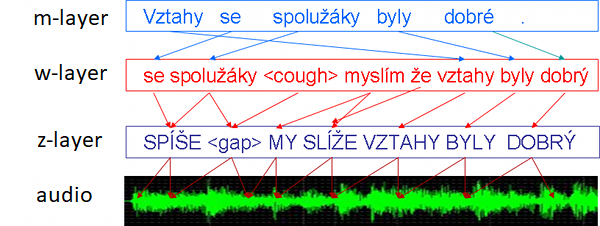}
\caption{ Annotation layers added for the spoken language dataset of PDT-C 1.0 
}
\label{fig:addedlayers}
\end{center}
\end{figure}

\section{Multi-layer Architecture}
\label{sec:layers}
The PDT-annotation scheme, described in detail in \newcite{haj17}, has a multi-layer architecture:
\begin{itemize}
\item \textbf{morphological layer} (m-layer): all tokens of the sentence get a lemma and morphological tag,
\item \textbf{surface syntactic layer} (a-layer): a dependency tree capturing surface syntactic relations such as subject, object, adverbial, etc.,
\item \textbf{deep syntactic layer} (t-layer) capturing the deep syntactic relations, ellipses, valency, topic-focus articulation, and coreference. In the process of the further development of the PDT, additional semantic features are being added to the original annotation scheme.
\end{itemize}

\vspace*{-7mm}
In addition to the above-mentioned three (main) annotation layers in the PDT-scenario, there is also the raw text layer (w-layer), where the text is segmented into documents and paragraphs and individual tokens are assigned unique identifiers. As it is mentioned in Sect.~\secref{sec:data}, there is additional audio and speech recognition layer (z-layer) in the spoken data. In the spoken data part (as opposed to the written corpora), the w-layer is in fact also an ``annotated'' layer, namely the manually provided transcription of the audio signal. 

\textbf{Linking the layers}. In order not to lose any piece of the original information, tokens (nodes) at a lower layer are explicitly referenced from the corresponding closest (immediately higher) layer. These links allow for tracing every unit of annotation all the way down to the original text, or to the transcript and audio (in the spoken data). 
It should be noted that while the inter-layer links are important 
for visualizing the trees and for training various NLP tools and applications, they are not part of any annotation layer from the theoretical point of view: the linguistic information should, be represented at the higher layer in its terms, without a loss.\footnote{An exception is the surface syntactic layer, where certain information from the morphology is missing (in the pure definition of its elements); consequently, the m-layer and a-layer are often taken together as a ``morphosyntactic''  annotation layer.} 

Figure~\ref{fig-layers} shows the relations between the neighboring layers as annotated and represented in the data. The rendered Czech sentence \textit{Byl by šel do lesa.} (‘lit.: He-was would went toforest.’) contains past conditional of the verb \textit{jít} (‘to go’) and a typo. These layers (from the w-layer to the t-layer) are part of all four datsets; the spoken dataset has, in addition, the audio and z-layers as depicted in Fig.~\ref{fig:addedlayers}.

In the following subsections, the manual annotation of the most important phenomena is shortly described.

\subsection{Spontaneous Speech Reconstruction}
Spontaneous speech reconstruction is a special type of manual annotation at the morphological layer that only belongs to the spoken data. The purpose of speech reconstruction is to ``translate'' the input "ungrammatical" spontaneous speech to a written text, before it is tagged and parsed. The transcript is segmented into sentence-like segments and these segments are edited to meet written-text standards, which means cleansing the text from the discourse-irrelevant and content-less material (e.g., superfluous connectives and deictic words, false starts, repetitions, etc. are removed) and re-chunking and re-building the original segments into grammatical sentences with acceptable word order and proper morphosyntactic relations between words. 
For more information about this type of manual annotation see \newcite{biblio:HaCiPDTSLAn2008} and \newcite{biblio:MiMiPDTSC202017}.

\subsection{Lemmatization and Tagging}
At the morphological layer, a lemma and a tag is assigned to each wordform. The annotation contains no syntactic structure, no attempt is even made to put together e.g. analytical verb forms or other types of multiword expressions. The annotation rules are described in  \newcite{novymanual}. 

There is manual annotation of lemmas and tags in all four datasets of PDT-C 1.0; for the new features, see Sect.~\secref{sec:morph}.

\subsection{Surface Syntactic Annotation}
 The surface syntactic annotation consists of dependency trees with surface syntactic function (dependency relation) assigned to every edge of the tree. A syntactic function determines the relation between the dependent node and its governing node (which is the node one level ``up'' the tree, in the standard visualization used in PDT-corpora). The annotation guidelines are described in  \newcite{analmanual}.\footnote{\url{http://ufal.ms.mff.cuni.cz/pdt2.0/doc/manuals/en/a-layer/html/index.html}} 
 
 For all the datasets in PDT-C 1.0, the surface syntactic annotation is the results of an automatic dependency parser except for the core written dataset which contains manual annotation of surface syntax, as it was done for its first released version PDT 1.0 \citelanguageresource{biblio:HaViPragueDependency2001}. Moreover, surface syntactic trees in the written data part are enriched with annotation of clause segmentation (cf. Table~\ref{tab:annot}), which was taken from the subsequent releases of PDT \cite{biblio:LoHoAnnotationsentence2012}. 
 
\subsection{Deep Syntactic Annotation}
One of the important distinctive features of the PDT-style annotation is the fact that in addition to the morphological and surface syntactic layer, it includes complex annotation of deep syntax, with certain semantic features, at the highest layer. Annotation principles used at the deep syntactic layer and the annotation guidelines are described in several annotation manuals \cite{trmanualanot,trmanualref,biblio:MiBeFromPDT2013,biblio:MiAnnotationtectogrammatical2014}.\footnote{\url{http://ufal.mff.cuni.cz/pdt2.0/doc/manuals/en/t-layer/html/index.html}}

At the deep syntactic layer, every sentence is represented as a rooted tree with labeled nodes and edges. The tree reflects the underlying dependency structure of the sentence. Unlike the lower layers, not all the original tokens are represented at this layer as nodes -- the nodes here stand for content words only.  Function words (prepositions, auxiliary verbs, etc.) do not have nodes of their own, but their contribution to the meaning of the sentence is not lost – several attributes are attached to the nodes the values of which represent such a contribution (e.g. tense for verbs). Some of the nodes do not correspond to any original token; they are added in case of surface deletions (ellipses). The types of the (semantic) dependency relations are represented by the \textit{functor} attribute attached to all nodes.

\subsection{Valency}
The core ingredient in the annotation of the deep syntactic layer is valency, the theoretical description of which, as developed in the framework of Functional Generative Description,  is summarized mainly in \newcite{panevova1974verbal}. The valency criterion divides functors into argument and adjunct functors. There are five core arguments: Actor ({\tt ACT}), Patient ({\tt PAT}), Addressee ({\tt ADDR}), Origin ({\tt ORIG}) and Effect ({\tt EFF}). In addition, about 50 types of adjuncts (temporal, local, casual, etc.) are used. For a particular verb (or more precisely, verb sense), a subset of the functors is obligatory, while others are either not present at all or are optional.

The valency lexicon that all the parts of PDT-C use, \textbf{PDT-Vallex}  \cite{biblio:HaPaPDTVALLEXCreating2003,biblio:UrBuildingPDTVALLEX2012}, was built in parallel with the manual annotation. It contains over 11,000 valency frames for more than 7,000 verbs which occurred in the datasets. It has been used for consistent annotation of valency: each occurrence of a verb in all corpora is linked to the appropriate valency frame in the lexicon.

\subsection{Coreference}
All parts of PDT-C except the user-generated data also capture grammatical and pronominal textual coreference relations (cf. Table~\ref{tab:annot}). Grammatical coreference is based on language-specific grammatical rules, whereas in order to resolve textual coreference, contextual knowledge is needed. Textual coreference annotation is based on the ``chain principle'', the anaphoric entity always referring to the last preceding coreferential antecedent. Coreference can be also cataphoric (point to the text that follows) and coreference links can span multiple sentences \cite{biblio:ZiHaDiscourseand2015}.

\subsection{Grammatemes}
So called grammatemes (described in detail in  \newcite{biblio:RaZaAnnotationGrammatemes2006}, \newcite{biblio:PaSeAnnotationMorphological2010}, \newcite{biblio:SePaGrammaticalnumber2010}) are attached to some nodes; they provide information about the node that cannot be derived from the deep syntactic structure, the functor and other attributes. Grammatemes are counterparts of those morphological categories which bear relevant deep syntactic or semantic information. 

Grammatemes are annotated at the deep syntactic layer only in the core PDT corpus (written texts; cf. Table~\ref{tab:annot}).

\subsection{Topic-Focus Articulation}
The information structure of the sentence (its topic-focus articulation) is expressed by various means (intonation, sentence structure, word order). 
It constitutes one of the basic aspects of the deep syntactic structure (for arguments on the semantic relevance of topic-focus articulation, see \newcite{SgallHP:1986}). The semantic basis of the articulation of the sentence into topic and focus is the relation of contextual boundness: a prototypical declarative sentence asserts that its focus holds (or does not hold, as the case may be) about its topic. Within both topic and focus, contextually-bound, contrastive contextually-bound, and non-bound nodes are distinguished \cite{biblio:HaSgTopicfocusarticulation1998}. 

The nodes at the deep structure layer are ordered according to the degrees of communicative dynamism.\footnote{The nodes at the surface-syntactic layer, as well as at the morphological layer, are of course naturally ordered based on the surface word order.}

In PDT-C 1.0, topic-focus articulation is captured at the deep syntactic layer only in the core PDT corpus (written text dataset; cf. Table~\ref{tab:annot}).

\subsection{Additional Annotation}
At the deep syntactic layer in the core PDT part (written dataset), valency of nouns, textual nominal coreference, bridging and discourse relations and other semantic properties of the sentence such as genre specification, multiword expressions, quotation are also annotated. More information of these special annotations can be found in general overview \cite{biblio:MiBeFromPDT2013}, and also in detailed studies \cite{biblio:LoHoAnnotationsentence2012,biblio:NeMiAnnotatingExtended2011,biblio:NeMiHowDependency2013,biblio:PoJiManualfor2012,biblio:BeStAnnotationMultiword2010,biblio:ZiHaDiscourseand2015}.


\section{New Annotation of Morphology}
\label{sec:morph}
As it has been already mentioned, the latest published versions of the included datasets have been enhanced in PDT-C 1.0 by a new or corrected manual annotation at the morphological layer (new in translated, spoken, and user-generated data, corrected in the original written dataset). Altogether, there are now 3,895,348 tokens with manual morphological annotation in the PDT-C 1.0 (cf. Table~\ref{tab:volume}).

\subsection{Annotation Process}

The annotation is based on a manual disambiguation of an automatic, dictionary-based morphological analysis of the annotated texts. For such automatic preprocessing, we use the MorphoDiTa tool \cite{strakova14}.\footnote{\url{https://ufal.mff.cuni.cz/morphodita}}

Key element to annotation consistency at the morphological layer is the  Czech morphological dictionary \textbf{MorfFlex} \citelanguageresource{morfflex}, which is now an integral part of the PDT-C 1.0 release.
The MorfFlex dictionary lists more than 100,000,000 lemma-tag-wordform triples. For each wordform, full inflectional information is coded in a positional tag. Wordforms are organized into entries (\textit{paradigm instances}, or \textit{paradigms} in short) according to their formal morphological behavior. The paradigm (set of wordforms) is identified by a unique lemma. The formal specification of the (original) dictionary is in \newcite{biblio:HaDisambiguationRich2004}. 

Based on the long-time experience with the usage of the dictionary and the current manual annotation of real data, we proposed to capture some phenomena differently in order to achieve better consistency within the dictionary as well as between the dictionary and the annotated corpora. The changes concern several complicated morphological features of Czech (a brief description of the changes is in the following subsections; for detailed information see \newcite{biblio:HlMiModificationsCzech2019} and \newcite{novymanual}).

In addition, we are enriching the MorfFlex dictionary with new words and wordforms found in the newly annotated texts. Some quantitative characteristics of the amended morphological dictionary and some statistical data about the changes made are tabulated in Tab.~\ref{tab:dict}. The number of paradigms that are now different from the original new version (i.e. there is a change in a lemma, comment attached to the lemma, form, and/or tag) is 299,055 paradigms; this means that 28.58\% of the dictionary has been modified.

\begin{table}[h]
\captionsetup{justification=centering}
\begin{center}
\begin{tabular}{l|r}
Description & Volume\\\hline
Paradigms in original version & 1,035,659\\
Paradigms in new version & 1,046,422 \\
Paradigms removed & 3,381 \\ 
Paradigms added & 14,144 \\ 
Paradigms changed & 299,055 \\ 
\end{tabular}
\caption{Statistics on the morphological dictionary and changes made for PDT-C 1.0}
\label{tab:dict}
 \end{center}
\end{table}

\textbf{Achieving consistency}. The changes in the dictionary have been projected back into the manually annotated data by repeated re-annotation to guarantee full consistency between all the data and the dictionary. 

While the sheer amount of annotated texts did not allow for multiple annotation (e.g., to determine inter-annotator agreement; in general, IAA on morphological annotation is about 97\%{} \cite{Hajic2004}) to finish in time with regard to the funding available, the consistency of annotation has been checked by a specific module using the information from the morphological dictionary and the annotated data. 

Within a detailed analysis of consistency between the data and dictionary, the following cases of inconsistencies are distinguished (they are ordered; a particular case applies only when none of the above cases apply):

\begin{itemize}
\item \textbf{Full matches}: An analysis of a wordfom (attached lemma and tag) in data fully matches an analysis in the dictionary.
\item \textbf{Unique lemma, comment change}: Compared to the analysis of a wordform in the data, there is a unique analysis in the dictionary that has same tag, same lemma (including lemma number) but differs in a comment (an additional descriptive element) attached to the lemma. 
\item \textbf{Unique lemma, sense change}: There is a unique analysis in the dictionary that has the same tag and the same lemma, but differs in the lemma index number. 
\item \textbf{Unique lemma, tag change}: There is a unique analysis in the dictionary that has same complete lemma (including lemma index number and comment), but differs in the tag. 
\item \textbf{Unique rest}: There is a unique analysis in the dictionary, and none of the variants above apply.
\item \textbf{Multiple lemma, comment change}: There are multiple analyses in the dictionary that have the same tag and the same lemma with an index number, but not the comment.
\item \textbf{Multiple lemma, sense change}: There are multiple analyses in the dictionary that have the same tag and the same lemma but differ in lemma number.
\item \textbf{Multiple lemma, tag change}: There are multiple analyses in the dictionary that have same complete lemma  but differ in tag. 
\item \textbf{Multiple rest}: There are multiple analyses in the dictionary, neither of the above variants apply.  
\item \textbf{No analysis}: There is no analysis in the dictionary for the given wordform.
\end{itemize}

An inconsistency between data and the dictionary (of all the types listed above) indicates an annotation problem or error in the dictionary. All inconsistencies have been corrected (mostly manually, partially  also automatically: e.g., changes in representation of verbal aspect (see below) have been done automatically, since the information has been found elsewhere). There are only full matches now (the first type in the above list), except for a small amount of wordform occurrences in the data that are not in the dictionary (but have manual analysis in the data); this applies mostly to foreign wordfoms and non-standard, sparse forms of Czech.

This now applies to all the four parts of PDT-C 1.0, including the almost 2 millions tokens in the translated, spoken and user-generated data that have been newly manually annotated for lemmas and tags, using the annotation and corrections procedure described herein. 


\begin{table}[h]
\captionsetup{justification=centering}
\begin{center}
\begin{tabular}{l|r|r}
Type of inconsistency & \% & Forms \\\hline
Full matches & 75.27\% & 1,473,162 \\
Unique lemma, comment change & 1.85\% & 36,297 \\
Unique lemma, sense change & 4.44\% & 86,977 \\
Unique, tag change & 7.19\% & 140,682 \\
Unique rest & 6.71\% & 131,347 \\
Multiple lemma, comment change & 0.00\% & 0 \\
Multiple lemma, sense change & 0.03\% & 591 \\
Multiple, tag change & 0.25\% & 4,803 \\
Multiple rest & 3.73\% & 72,924 \\
No analysis & 0.53\% & 10,459 \\
\end{tabular}
\caption{Analysis of inconsistencies between the original PDT-data and the new version of the MorfFlex dictionary}
\label{tab:consistency}
\end{center}
\end{table} 

In the written dataset, i.e. in the original data of PDT (where the manual morphological annotation has already been completed for the original PDT 1.0 version \citelanguageresource{biblio:HaViPragueDependency2001}),  the annotation at the morphological layer has been checked and corrected to reflect the updated morphological annotation guidelines and also to be fully consistent with the new morphological dictionary. Tab.~\ref{tab:consistency} quantifies the amount of inconsistencies between the originally annotated data and the new version of the dictionary. All mismatches (25\% tokens) have been resolved. 


\subsection{New Features in Lemmatization}

The \textit{lemma} is a unique identifier of the paradigm. Usually it is the base form of the word (e.g. infinitive for verbs, nominative singular for nouns), possibly followed by a number distinguishing different lemmas with the same spelling. 

\textbf{Lemma numbering} has been improved and made more consistent. Now, we do not strive to make any distinction between meanings of homonyms. The only differences we want to capture are those of formal morphological nature. It means that we add numbers only to lemmas that have:
\begin{itemize}
\item different POS, e.g. \textit{růst-1} as noun (‘a growth’) and \textit{růst-2} as verb (‘to grow’),
\item different gender/declension for nouns, e.g. \textit{kredenc-1} as masculine and \textit{kredenc-2} as feminine, even if they have the same meaning (‘a cupboard’),
\item different aspect and/or conjugation in case of verbs, e.g. \textit{stát-1} with perfective aspect (‘to happen’) and \textit{stát-2} with imperfective aspect (‘to stand’).
\end{itemize}

Thus, we have, e.g., lemma \textit{jeřáb-1} for crane as a bird (animate masculine) and \textit{jeřáb-2} for both a tree and crane as a device for lifting heavy objects (inanimate masculine). We do not distinguish the latter two meanings (tree vs. device), because they do not differ from the inflectional point of view (same declension). There might be a difference in derivation; e.g. the word \textit{jeřábník} (a man who works with a crane-device) is derived from \textit{jeřáb} as a device. This has been delegated to derivational data sources, such as \newcite{vidra-etal-2019-derinet}; here, we do not to take such derivational, stylistic and semantic differences into account.

\textbf{Orthographic and stylistic variants} of a word (e.g., an archaic variant \textit{these}, a standard variant \textit{teze}, and a non-standard variant \textit{téze} ‘thesis’) were not tackled uniformly in MorfFlex. Some variants had different paradigms with different lemmas, others were grouped into one paradigm with one common lemma. In the former case there was no connection between the two variant lemmas. The latter case led to the most massive violations of the principles of the dictionary because there were different wordforms with the same tags belonging to the same lemma. 
We have decided to capture variants in separate paradigms with different lemmas: we select one of the variants as ``basic'' (the standard one, i.e. \textit{teze}) and other variants (non-standard \textit{téze} and archaic \textit{these}) refer to it in an additional descriptive element, attached to the lemma. We have introduced new codes for marking variants of different style; cf. lemmas of variants of word \textit{teze} in Table~\ref{tab:variants}.

\begin{table}[h]
\captionsetup{justification=centering}
\begin{center}
\begin{tabular}{l|l}
Wordform & Lemma\\\hline
{\it teze} & teze\\
{\it these} & these\_,a\_$^\wedge$($^\wedge$DD**teze)\\
{\it téze} & téze\_,h\_$^\wedge$($^\wedge$GC**teze)\\
\end{tabular}
\caption{Capturing stylistic variants: word \textit{teze} ‘thesis’}
\label{tab:variants}
 \end{center}
\end{table}

\vspace*{-7mm}

\begin{table}[h]
\captionsetup{justification=centering}
\begin{center}
\begin{tabular}{r|l}
Position & Description \\\hline
\textbf{1} & \textbf{Part of speech} \\
\textbf{2} & \textbf{Detailed part of speech} \\
3 & Gender \\
4 & Number \\
5 & Case \\
6 & Possessor's gender \\
7 & Possessor's number \\
8 & Person \\
9 & Tense \\
10 & Degree of comparison \\
11 & Negation \\
12 & Voice \\
\textbf{13} & \textbf{Verbal aspect} \\
\textbf{14} & \textbf{Aggregate} \\
\textbf{15} & \textbf{Variant, style, abbreviation}
\end{tabular}
\caption{Attributes in positional tags (amended positions are in bold font)}
\label{tab:tag}
\end{center}
\end{table} 

\begin{table*}[h]
\begin{center}
\begin{tabular}{l|l|l|l}
Wordform & Lemma & Tag & Example \\
\hline
{\it Wall} & {\tt Wall} & \texttt{\underline{F\%{}}-------------} &  {\it na Wall Street}  ‘on the Wall Street’ \\
{\it česko} & {\tt česko} & \texttt{\underline{S2}--------A----} &  {\it česko-ruská kniha}  ‘Czech-Russian book’ \\
{\it kou} & {\tt ka} & \texttt{\underline{SN}FS7-----A----} &  {\it s manželem/kou}   ‘with husband/wife’ \\
{\it připravil} & {\tt připravit} & \texttt{VpYS----R-AA\underline{P}--} &  {\it Připravil návrh.} '[He] has prepared a proposal.' \\
{\it proň} & {\tt on} & \texttt{P5ZS4--3-----\underline{p}-} &  {\it Žije proň.}  ‘[He] lives for him.’ \\
{\it s} & {\tt strana} & \texttt{NNFXX-----A---\underline{a}} &  {\it na s. 12}  ‘at page 12’ \\
\end{tabular}
\caption{Examples of annotation (amended positions/values are underlined)}
\label{tab:examplesnewtags}
\end{center}
\end{table*} 

\vspace*{-6mm}
\subsection{New Features in Tagging}
Czech is a highly inflectional language. There are 15 categories, encoded in a positional tag, which is a string of 15 characters; every position encodes one morphological category using one character symbol. An overview of the 15 positions is in Table~\ref{tab:tag}. The categories we have newly defined and/or amended are indicated in bold. Examples of the use of the modified categories are shown in Table~\ref{tab:examplesnewtags}.\footnote{For the original full list of tag categories and values, see e.g. \url{https://ufal.mff.cuni.cz/pdt/Morphology_and_Tagging/Doc/hmptagqr.html}.}

\textbf{Foreign words}. Two new POS have been introduced: for foreign words and for segments (see below). Their detailed description can be found in \cite{biblio-7044017271673886902}. Foreign word is such word that is not subject to Czech inflectional system and has no meaning of its own in Czech. The tag contains special values at the POS and detailed POS position, namely {\tt F\%}. There are no other morphological values involved in the tag (cf. the example of wordform \textit{Wall} in Table~\ref{tab:examplesnewtags}). 

\textbf{Segments} are incomplete words. In order to understand them, they must be joined with another string or word to create a complete word. 
We have created a new POS with the code {\tt S} for them. According to their position in the complete word, we distinguish prefixal and suffixal segments. The tag of the prefixal segments has the code {\tt 2} at the 2nd position (cf. the wordform \textit{česko} in Table~\ref{tab:examplesnewtags}). The suffixal segments express an affiliation to a specific POS. Thus, all the inflectional categories that describe the whole wordform, except for the first one (which is {\tt S}), are filled in the tag (cf. wordform \textit{kou} in Table~\ref{tab:examplesnewtags}). 

\textbf{Aspect}. The verbal aspect was not part of the tag, the information was included in the dictionary in a form of an additional field attached to lemmas. We have added the information about the aspect directly to the tag, to its 13th position, which had been kept free as a reserve. The values are: {\tt P} for perfective verbs, {\tt I} for imperfective ones and {\tt B} for verbs with both aspects. There is an example of verbal wordform \textit{připravil} in Table~\ref{tab:examplesnewtags}.

\textbf{Aggregates}. New solution has also been implemented for so-called aggregates. An aggregate is a wordform created by combining two or more forms (components of the aggregate) into one and cannot be simply assigned any POS (e.g. wordform \textit{proň} consists of a pronoun \textit{on} (‘he’) and the joined preposition \textit{pro} (‘for’)). The tag describes the main component of the aggregate (i.e the pronoun \textit{on}) 
and the joined components are coded at the free 14th position of the tag (for joined preposition \textit{pro}, there is value {\tt p}; cf. wordform \textit{proň} in Table~\ref{tab:examplesnewtags}).

\textbf{Stylistic variants, abbreviations}. For marking stylistic variants of a wordform (e.g. both \textit{orli} and \textit{orlové} (‘eagles’) are the wordforms of the noun \textit{orel} (‘eagle’) and express plural masculine nominative), we use the 15th position of the tag, as has been done before. The main difference lies in the fact that now we use this position strictly for variants of wordform. 
Numbers 1 to 4 mark standard variants, while numbers 5 to 9 relate to substandard ones. We have also added new values -- letter {\tt a}, {\tt b} and {\tt c} -- to the 15th position of the tag for marking abbreviation of a (single) word which is captured as a special wordform of the paradigm of that word (cf. abbreviation \textit{s} (‘p’) which abbreviates word \textit{strana} (‘page’) in Table~\ref{tab:examplesnewtags}). 

\section{Conclusion and Future Work}

A large, combined genre-diversified treebank resource for Czech with enhanced morphological annotation has been presented, which increases the amount of morphologically annotated data for Czech almost twofold, to nearly 4 million tokens. PDT-C 1.0 will be published under an open license in the first months of 2020, together with the morphological and valency lexicons related to the annotation. A large number of changes has been made not only to the data, but also to the morphological dictionary MorfFlex, and consistency is now assured across all of the four original datasets. 

In the near future, the combined corpus will be used for building a new model for Czech morphological disambiguation tool MorphoDiTa, which will in turn be used for automatic POS and morphological annotation of all the Czech corpora available in the Kontext KWIC tool at the LINDAT/CLARIAH-CZ research infrastructure.\footnote{\url{https://clariah.lindat.cz}} It will also be made available again for the Czech National Corpus\footnote{\url{https://www.korpus.cz}} to re-annotate its corpora with a more accurate POS and morphology. The whole PDT-C 1.0 will also be available for advanced search in the PML-TQ tool.\footnote{\url{https://lindat.mff.cuni.cz/services/pmltq}}


On the annotation side, surface syntactic annotation of the three so far manually unannotated datasets from the PDT-C 1.0 will be tackled next, again in order to increase the amount of data available to train dependency parsers for Czech. Having both morphology as well as dependency syntax annotated will then allow to increase the amount of Czech data in the Universal Dependencies \cite{NIVRE16.348} collection to almost 5 million; the conversion to the UD style of annotation will be straightforward, as it was already in the case of PDT \cite{biblio-1745977273001647149}.

\section*{Acknowledgements}
The research and language resource work reported in the paper has been supported by the LINDAT/CLARIN and LINDAT/CLARIAH-CZ projects funded by Ministry of Education, Youth and Sports of the Czech Republic (projects LM2015071, LM2018101 and EF16\_{}013/0001781). The original annotation has been supported by multiple projects in the past, funded both nationally by the Ministry of Education, Youth and Sports of the Czech Republic and the Czech Science Foundation, such as Projects No. VS96151, LN00A063, LC536, LM2010013, GA405/96/0198,  GV405/96/K214), as well as projects funded by the European Commission in the 6th and 7th Framework Programmes and the H2020 Programme that in part added certain language resources, such as the Companions and Khresmoi Integrated Projects, the Faust STREP and several others.

\section*{Bibliographical References}
\label{main:ref}
\bibliographystyle{lrec}
\bibliography{biblio}

\section*{Language Resource References}
\label{lr:ref}
\bibliographystylelanguageresource{lrec}
\bibliographylanguageresource{biblio}

\end{document}